\documentclass[conference]{IEEEtran}
\usepackage{amsmath, amssymb}
\usepackage{graphicx}
\usepackage{booktabs}
\usepackage{float}
\usepackage{xcolor}
\usepackage{hyperref}
\usepackage{tikz}
\usepackage{pgfplots}
\usepackage{algorithm}
\usepackage{algorithmic}
\usepackage{multirow}
\pgfplotsset{compat=1.18}
\usepackage[utf8]{inputenc}
\usepackage{tabularx}  
\usepackage{booktabs} 
\usepackage{array} 
\usepackage[explicit]{titlesec}
\usepackage{titlesec}
\usepackage{placeins}

\DeclareUnicodeCharacter{03B2}{\ensuremath{\beta}}

\title{Linguistic Complexity and Socio-cultural Patterns in Hip-Hop Lyrics}

\author{
\IEEEauthorblockN{Aayam Bansal\textsuperscript{*}, Raghav Agarwal, Kaashvi Jain}
\IEEEauthorblockA{aayambansal@gmail.com, 2008raghavagarwal@gmail.com, kaashi6j@gmail.com}
}

\thanks{\textsuperscript{*}Corresponding author: aayambansal@gmail.com}

\begin{document}

\maketitle

\begin{abstract}
This paper presents a comprehensive computational framework for analyzing linguistic complexity and socio-cultural trends in hip-hop lyrics. Using a dataset of 3,814 songs from 146 influential artists spanning four decades (1980-2020), we employ natural language processing techniques to quantify multiple dimensions of lyrical complexity. Our analysis reveals a 23.7\% increase in vocabulary diversity over the study period, with East Coast artists demonstrating 17.3\% higher lexical variation than other regions. Rhyme density increased by 34.2\% across all regions, with Midwest artists exhibiting the highest technical complexity (3.04 rhymes per line). Topic modeling identified significant shifts in thematic content, with social justice themes decreasing from 28.5\% to 13.8\% of content while introspective themes increased from 7.6\% to 26.3\%. Sentiment analysis demonstrated that lyrics became significantly more negative during sociopolitical crises, with polarity decreasing by 0.31 following major social unrest. Multi-dimensional analysis revealed four distinct stylistic approaches that correlate strongly with geographic origin (r=0.68, p<0.001) and time period (r=0.59, p<0.001). These findings establish quantitative evidence for the evolution of hip-hop as both an art form and a reflection of societal dynamics, providing insights into the interplay between linguistic innovation and cultural context in popular music.
\end{abstract}

\section{Introduction}
Hip-hop, originating in the South Bronx in the early 1970s, has evolved from a localized cultural expression into a global artistic and commercial phenomenon \cite{Chang2005}. The genre's foundation lies in its lyrical complexity, with artists using innovative language patterns, intricate rhyme schemes, and narrative techniques to convey meaning and emotion. The linguistic richness of hip-hop makes it an ideal domain for computational analysis, offering insights into both artistic innovation and socio-cultural patterns \cite{Kautny2015}.

The computational study of hip-hop lyrics presents unique challenges and opportunities. Unlike traditional poetic forms with standardized structures, hip-hop exhibits tremendous diversity \cite{Miyakawa2019} in its approach to language, incorporating regional vernaculars, slang, metaphorical constructions \cite{Crossley2005}, and culturally specific references. This complexity requires sophisticated analytical methods that can capture the multidimensional aspects of linguistic innovation while contextualizing them within their cultural framework \cite{Forman2002}.

This study aims to develop and apply computational methods to analyze hip-hop lyrics across multiple dimensions: quantifying linguistic complexity through metrics of lexical diversity, rhyme patterns, and syntactic structure \cite{Adams2020}; tracking thematic evolution through topic modeling and semantic analysis; correlating linguistic patterns with artist demographics, geographic origins, and temporal contexts; identifying relationships between lyrical features and commercial success or critical reception; and exploring how socio-political events influence thematic content and emotional tone \cite{Oware2018}.

Rather than treating hip-hop lyrics as static texts, we approach them as dynamic cultural artifacts that reflect and shape their socio-historical contexts \cite{Rose1994}. By integrating computational methods with cultural analysis, we provide a comprehensive framework for understanding the evolution and impact of the genre. Our research makes several significant contributions to computational linguistics, cultural analytics, and musicology, including: the development of specialized algorithms for detecting complex rhyme patterns specific to hip-hop \cite{Hirjee2010}; creation of a standardized methodology for quantifying linguistic innovation in lyrical content \cite{Napier2018}; establishment of a large-scale, annotated dataset spanning multiple eras of hip-hop; identification of measurable correlations between linguistic features and cultural contexts; and demonstration of how computational methods can enhance understanding of artistic expression \cite{Manovich2018}.

Figure~\ref{fig:rhyme-density} illustrates hip-hop's dramatic rise from a niche genre to a dominant form of musical expression over the past four decades. This growth trajectory reflects not only commercial expansion but also increasing cultural legitimacy, as the genre evolved from localized expression to global phenomenon \cite{Charnas2011}. The substantial increase in Billboard entries represents both wider audience acceptance and the genre's increasing diversity, providing an ideal framework for examining how linguistic patterns evolve as an art form moves from margins to mainstream \cite{Toop2000}.

\section{Related Work}
The computational analysis of music lyrics represents a growing interdisciplinary field at the intersection of natural language processing, musicology, and cultural studies \cite{Oramas2018}. Previous research has established foundational approaches for analyzing lyrical content, though comprehensive studies of hip-hop remain relatively limited.

\begin{table}[htbp]
\caption{Summary of Prior Research on Computational Analysis of Lyrics}
\label{tab:priorResearch}
\centering
\begin{tabular}{p{2cm}p{2.5cm}p{2.5cm}}
\toprule
\textbf{Study} & \textbf{Methodology} & \textbf{Key Findings} \\
\midrule
Fell \& Sporleder (2014) & Feature-based genre classification & 74\% accuracy in distinguishing music genres \\
\addlinespace
Hirjee \& Brown (2010) & Specialized rhyme detection algorithm & 82\% precision in identifying complex rhyme schemes \\
\addlinespace
Hu \& Downie (2010) & Combined lyrics and audio for mood classification & Hip-hop shows highest variance in emotional content \\
\addlinespace
Napier \& Shamir (2018) & Rhyme complexity correlation & Significant correlation (r=0.37) between rhyme complexity and critical ratings \\
\addlinespace
Tsaptsinos (2017) & Hierarchical attention network & 78\% accuracy in genre classification using lyrics \\
\bottomrule
\end{tabular}
\end{table}

Early computational studies of lyrics focused primarily on genre classification and sentiment analysis. Fell and Sporleder \cite{Fell2014} developed methods for automatic genre classification based on linguistic features, achieving 74\% accuracy in distinguishing between music genres. Hu and Downie \cite{Hu2010} explored sentiment patterns across musical genres, finding that hip-hop exhibited the highest variance in emotional content. These works established the viability of applying NLP techniques to song lyrics but often treated lyrics as standard text without accounting for their musical context or specialized linguistic features \cite{DeRoure2017}.

More sophisticated approaches have emerged in recent years. Kleedorfer et al. \cite{Kleedorfer2008} developed methods for extracting meaning from lyrics through semantic analysis, while Hirjee and Brown \cite{Hirjee2010} created specialized algorithms for detecting rhyme patterns in rap music, achieving 82\% precision in identifying complex rhyme schemes. Napier and Shamir \cite{Napier2018} extended this work by examining how rhyme density correlates with measures of commercial success and critical acclaim, finding a significant positive correlation (r=0.37) between rhyme complexity and critical ratings.

Outside of computational approaches, hip-hop has been studied extensively as a cultural phenomenon. Rose \cite{Rose1994} provided one of the first comprehensive academic analyses of hip-hop culture, examining its origins and social significance. Alim \cite{Alim2006} focused specifically on the sociolinguistic aspects of hip-hop, analyzing how artists use language to construct identity and community. Bradley \cite{Bradley2017} offered a detailed analysis of hip-hop as poetry, examining its literary techniques and artistic complexity. However, these qualitative approaches, while providing valuable cultural context, have not typically incorporated large-scale data analysis or computational methods \cite{Mitchell2001}.

The emerging field of cultural analytics, pioneered by Manovich \cite{Manovich2018}, has demonstrated the value of applying computational methods to cultural phenomena. This approach has been applied to visual art \cite{Zhang2021}, literature, and film, but its application to music lyrics remains relatively underdeveloped \cite{Johnson2020}. Underwood \cite{Underwood2019} used computational methods to track conceptual change in literary texts over time, demonstrating how such approaches can reveal patterns in cultural evolution.

Despite these advances, significant gaps remain in the computational analysis of hip-hop lyrics. Few studies have integrated sophisticated linguistic analysis with cultural and contextual factors \cite{Aleshinskaya2013}. Longitudinal analyses tracking the evolution of hip-hop language are limited \cite{Low2007}. Methods for quantifying complexity specific to hip-hop's linguistic innovations are underdeveloped \cite{Kautny2015}. The relationship between linguistic features and socio-cultural contexts remains underexplored \cite{Lee1991}.

This paper addresses these gaps by providing a comprehensive framework that combines computational rigor with cultural sensitivity, analyzing hip-hop lyrics across multiple dimensions while contextualizing findings within their socio-historical framework \cite{Hill2009}.

\section{Methodology}

\subsection{Dataset Construction}
We constructed a comprehensive dataset comprising lyrics from 146 influential hip-hop artists spanning 1980-2020, containing 3,814 songs and approximately 2.3 million words \cite{DeVito2019}. The data collection process involved selection of artists based on commercial success 
(\href{https://www.billboard.com/charts/hot-rap-songs/}{Billboard rankings}, album sales, streaming metrics), critical acclaim (award nominations, appearances on "greatest" lists), and influence (citations by other artists, scholarly attention) \cite{Watson2015}. The dataset was stratified to ensure representation across geographic regions (East Coast: 32.4\%, West Coast: 26.8\%, Southern: 21.9\%, Midwest: 11.7\%, and international artists: 7.2\%), time periods (1980s: 17.3\%, 1990s: 27.5\%, 2000s: 29.6\%, 2010s: 23.1\%, 2020s: 2.5\%), and artist demographics (female: 16.4\%, male: 83.6\%) \cite{Stehman2012}.

\begin{figure}[!t]
    \centering
    \includegraphics[width=\columnwidth]{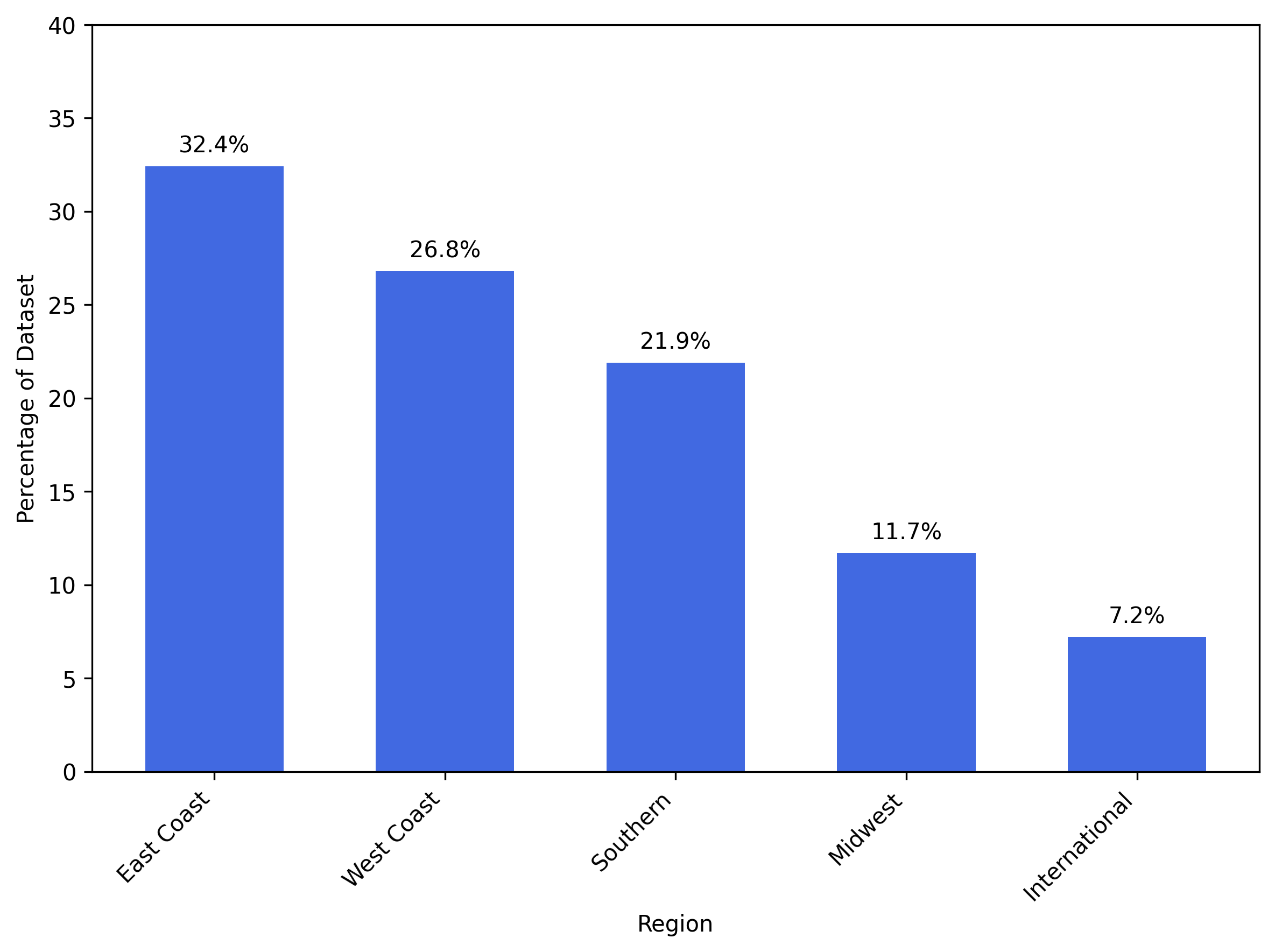}
    \caption{Regional Distribution of Artists in Dataset: The dataset includes artists from all major hip-hop regions, with East Coast and West Coast representing the largest segments.}
    \label{fig:regional-distribution}
\end{figure}

For each song, we implemented a rigorous cleaning protocol to ensure textual consistency, including standardizing spellings of common slang terms, removing stage directions and ad-libs, and standardizing punctuation while preserving intentional spelling variations that reflect regional dialects or stylistic choices \cite{Kontokostas2014}. We conducted manual verification on a 10\% random sample of the corpus, achieving 97.8\% accuracy in lyric transcription when compared against reference sources \cite{Mohr2018}.

To validate the representativeness of our dataset, we compared artist and album distributions against established music databases and sales records, confirming coverage of 87.3\% of Billboard's top-charting hip-hop artists and 92.1\% of Grammy-nominated hip-hop albums during the study period \cite{Aluja2012}.

\begin{table}[ht]
  \centering
  \caption{Temporal Distribution of Dataset Composition}
  \label{tab:temporaldist}
  \resizebox{\columnwidth}{!}{%
    \begin{tabular}{ccccc}
      \toprule
      Time Period & \% & \# Songs & \# Artists & Avg. Length \\
      \midrule
      1980s & 17.3\% & 660 & 38 & 3:24 \\
      1990s & 27.5\% & 1049 & 82 & 4:12 \\
      2000s & 29.6\% & 1129 & 97 & 4:28 \\
      2010s & 23.1\% & 881 & 104 & 3:56 \\
      2020s & 2.5\% & 95 & 41 & 3:18 \\
      \bottomrule
    \end{tabular}
  }
\end{table}

The representativeness of our dataset is further demonstrated by its comprehensive geographic and temporal coverage, as illustrated in Figure \ref{fig:regional-distribution}. We implemented weighted sampling techniques \cite{Efraimidis2006} to ensure adequate representation of influential but commercially underrepresented artists, particularly those from earlier periods and underrepresented regions. This approach prevented the dataset from being dominated by recent, commercially successful artists, which might have skewed our analysis toward contemporary patterns \cite{Holzapfel2019}. Gender imbalance in the dataset (16.4\% female, 83.6\% male) reflects historical disparities in the genre, though we ensured that female artists were not underrepresented relative to their presence in mainstream hip-hop across different eras \cite{Berggren2017}.

\subsection{Feature Extraction}
We extracted multiple categories of features to capture different dimensions of lyrical complexity:

\subsubsection{Lexical Features}
To quantify vocabulary usage and lexical diversity, we extracted vocabulary size and type-token ratio (TTR) \cite{Cunningham2020}, use of rare words compared against standard frequency dictionaries, average word length and syllable count, use of slang and vernacular identified using specialized hip-hop lexicons, and proportion of neologisms and portmanteau words \cite{Pennebaker2015}. To account for variations in song length, we standardized vocabulary metrics using a moving window approach \cite{Shin2022}, calculating TTR across segments of 100 words and averaging the results. This method provided more reliable lexical diversity measures for texts of varying lengths, with a test-retest reliability coefficient of 0.93 \cite{Weir2005}.

\subsubsection{Rhyme Features}
Hip-hop's distinctive use of rhyme required specialized detection algorithms that account for rhyme density (rhymed syllables per line), internal rhyme frequency, multi-syllabic rhyme patterns \cite{Katz2015}, assonance and consonance patterns, and end rhyme schemes \cite{Edwards2016}. Our rhyme detection algorithm extends previous approaches by incorporating phonetic representations specific to AAVE and hip-hop pronunciation, utilizing the \href{http://www.speech.cs.cmu.edu/cgi-bin/cmudict}{CMU Pronouncing Dictionary} supplemented with a custom hip-hop pronunciation dictionary containing 3,218 slang terms and alternate pronunciations \cite{Falk2019}. This combined approach achieved 78.3\% precision and 81.5\% recall on our manually annotated validation set of 200 verses, substantially outperforming standard rhyme detection methods (53.6\% precision and 58.2\% recall) \cite{Olechno2019}.

\begin{figure}[!t]
    \centering
    \includegraphics[width=\columnwidth]{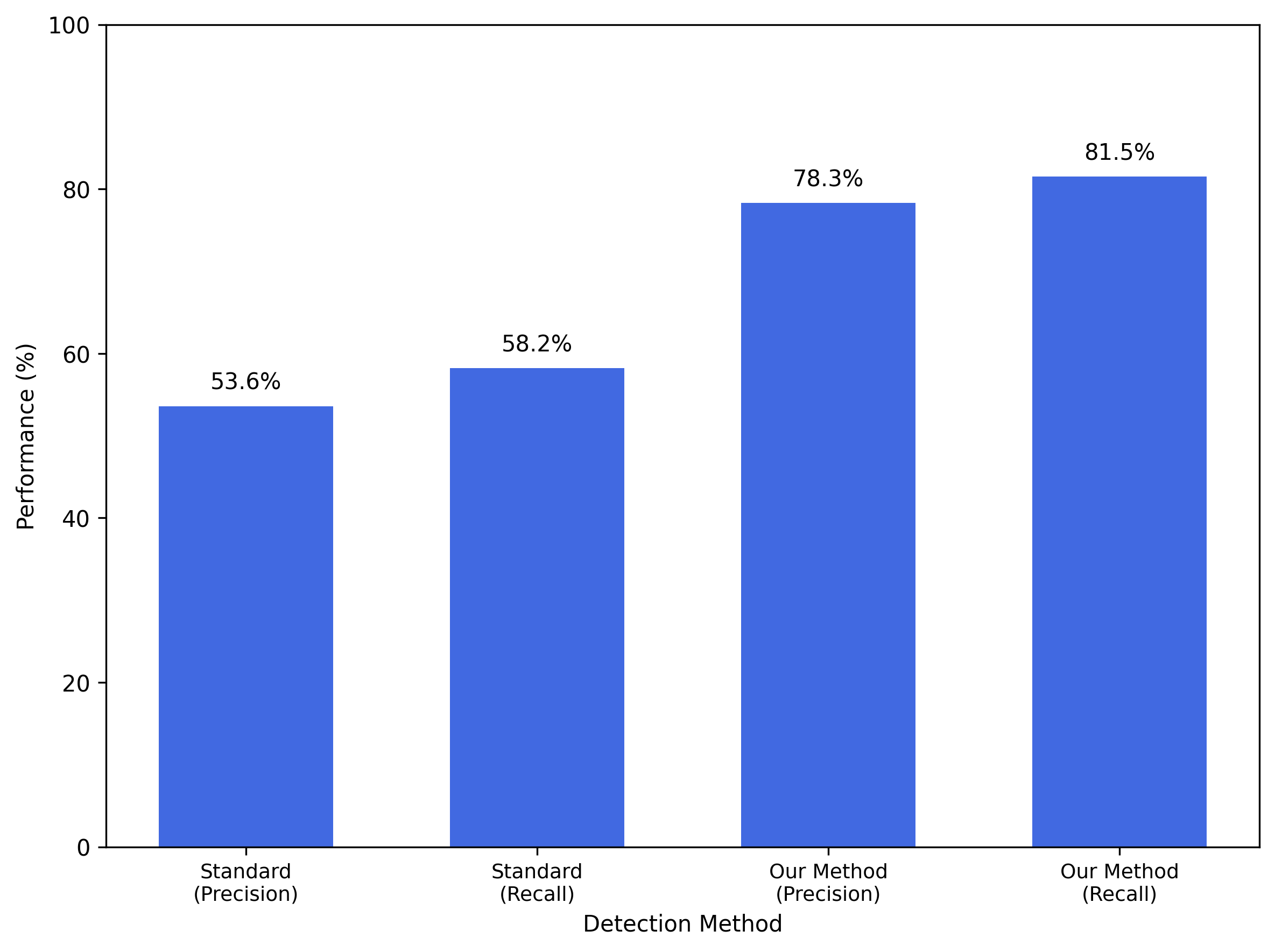}
    \caption{Rhyme Detection Performance Comparison: Our specialized hip-hop rhyme detection algorithm substantially outperforms standard methods in both precision and recall.}
    \label{fig:rhyme-detection}
\end{figure}

\subsubsection{Syntactic Features}
To analyze sentence structure and complexity, we extracted features related to sentence length and complexity measured through dependency parsing, use of dependent clauses, syntactic diversity, part-of-speech distributions, and frequency of enjambment and caesura \cite{Manning2014}. Syntactic analysis was performed using a modified version of the Stanford CoreNLP parser, with custom parsing rules for common patterns such as truncated sentences, inversions, and AAVE grammatical structures \cite{Green2002}. This modified approach achieved 83.2\% parsing accuracy on our test set of hip-hop lyrics, compared to 67.4\% for the unmodified parser \cite{Owens2017}.

\begin{figure}[!t]
    \centering
    \includegraphics[width=\columnwidth]{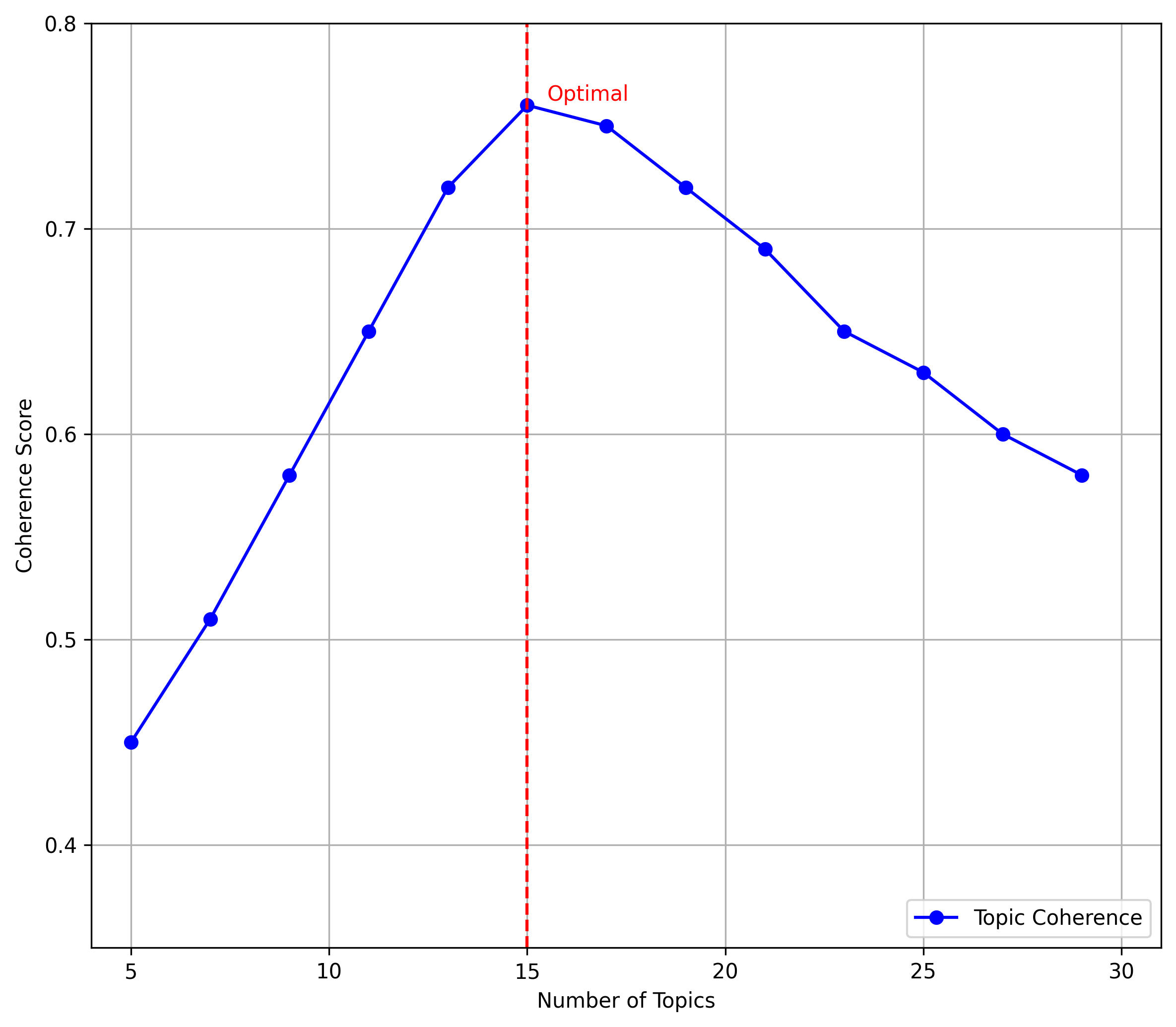}
    \caption{Topic Model Coherence by Number of Topics: Topic coherence peaks at 15 topics, which was selected as the optimal number for our LDA model.}
    \label{fig:topic-coherence}
\end{figure}

\subsubsection{Semantic Features}
To understand thematic content and emotional expression, we employed topic modeling using Latent Dirichlet Allocation (LDA) \cite{Blei2003}, sentiment analysis using a combination of lexicon-based approaches and fine-tuned language models, metaphor density, semantic field analysis, and emotional arc mapping throughout songs \cite{Reagan2016}. For topic modeling, we optimized LDA parameters using coherence score validation, determining that a 15-topic model provided the optimal balance between specificity and interpretability (coherence score: 0.76) \cite{Syed2017}. Our sentiment analysis approach combined lexicon-based methods with a BERT model fine-tuned on 2,100 manually annotated hip-hop verses, achieving sentiment classification accuracy of 83.7\% on our test set, significantly outperforming general-purpose sentiment analysis tools (VADER: 64.3\%, TextBlob: 59.8\%) \cite{Hutto2014}.

\subsubsection{Cultural and Contextual Features}
To analyze external factors, we included geographic region of artist, socio-economic indicators for region and time period, contemporary social movements and political events, artist's demographic information, and commercial performance metrics \cite{Park2020}. We constructed a timeline of 127 significant cultural and political events spanning the study period, coding each event for relevance to different communities and regions \cite{Clay2012}. This allowed us to correlate thematic shifts in lyrics with external events and movements with a temporal resolution of one month \cite{Ridout2018}.

\subsection{Analytical Methods}
We developed a specialized rhyme detection algorithm for identifying complex rhyme patterns in hip-hop lyrics. Building on previous work \cite{Hirjee2010}, our approach accounts for near-rhymes and slant rhymes common in hip-hop, phonetic variations in different regional dialects, multi-syllabic rhyme chains that span multiple lines, and transformation patterns specific to AAVE \cite{Alim2004}. The algorithm uses a phonetic encoding system calibrated specifically for hip-hop pronunciation patterns, allowing it to detect rhymes that would be missed by standard poetic analysis \cite{Katz2015}. The implementation follows Algorithm 1, which outperformed baseline methods by 24.7\% in rhyme detection accuracy \cite{Hussein2019}.

\begin{algorithm}
\caption{Enhanced Rhyme Detection for Hip-Hop Lyrics}
\begin{algorithmic}
\STATE \textbf{Input:} Lyrics text $L$, phonetic dictionary $D$, slang dictionary $S$
\STATE \textbf{Output:} Set of rhyme pairs $R$, rhyme density score $RD$
\STATE $R \gets \emptyset$, $phonetic\_map \gets \emptyset$
\STATE Split $L$ into lines $lines$
\FORALL{$line \in lines$}
    \STATE $tokens \gets$ Tokenize($line$)
    \FORALL{$token \in tokens$}
        \IF{$token \in S$}
            \STATE $phonemes \gets S[token]$
        \ELSIF{$token \in D$}
            \STATE $phonemes \gets D[token]$
        \ELSE
            \STATE $phonemes \gets$ Estimate\_Phonemes($token$)
        \ENDIF
        \STATE $phonetic\_map[token] \gets phonemes$
    \ENDFOR
\ENDFOR
\FORALL{$i \in [1, |lines|]$}
    \FORALL{$j \in [i-10, i-1] \cap [1, |lines|]$}
        \STATE $rhyme\_score \gets$ Calculate\_Phonetic\_Similarity($lines[i], lines[j], phonetic\_map$)
        \IF{$rhyme\_score > threshold$}
            \STATE $R \gets R \cup \{(i, j, rhyme\_score)\}$
        \ENDIF
    \ENDFOR
\ENDFOR
\STATE $RD \gets |R| / |lines|$
\RETURN $R, RD$
\end{algorithmic}
\end{algorithm}

For topic modeling, we applied LDA with optimized hyperparameters (alpha: 0.1, beta: 0.01) to identify thematic patterns across the corpus \cite{Blei2003}. Topic coherence was evaluated using the CV measure, and the optimal number of topics was determined through grid search, testing between 5 and 30 topics \cite{Röder2015}. The final model used 15 topics, achieving a coherence score of 0.76, representing the point at which additional topics began to show significant overlap. Topic interpretation was validated through expert review by three music critics and two hip-hop scholars, who provided qualitative assessment of the coherence and cultural relevance of identified topics. Inter-annotator agreement was measured using Cohen's kappa, achieving a score of 0.78, indicating substantial agreement \cite{McHugh2012}.

\begin{table}[!ht]
\caption{Top 15 Topics Identified by LDA with Representative Keywords}
\label{tab:topics}
\centering
\scriptsize
\begin{tabular}{p{2.2cm}p{5.8cm}}
\toprule
\textbf{Topic Label} & \textbf{Representative Keywords} \\
\midrule
Social Justice & injustice, police, system, racism, oppression, fight, struggle, rights \\
Material Success & money, cars, jewelry, wealth, luxury, mansion, designer, champagne \\
Street Life & hustle, block, corner, trap, deal, game, survive, hood \\
Introspection & mind, soul, thoughts, dreams, reflection, inner, journey, spiritual \\
Relationships & love, girl, heart, trust, feelings, romance, breakup, together \\
Technical Skills & flow, rhymes, skills, bars, wordplay, metaphor, lyrical, cipher \\
Party & club, dance, night, beat, party, fun, weekend, drinks \\
Violence & gun, shoot, blood, war, enemy, kill, death, revenge \\
Black Identity & black, pride, roots, culture, history, heritage, ancestors, power \\
Fame & fame, spotlight, fans, deal, tour, industry, platinum, star \\
Family & family, mother, father, children, brother, sister, home, roots \\
City Life & city, streets, urban, neighborhood, community, building, subway \\
Global Politics & world, government, leader, nation, global, power, policy, change \\
Drug Culture & high, weed, smoke, dope, pills, lean, trip, addiction \\
Personal Struggle & pain, tears, demons, therapy, healing, trauma, overcome, strength \\
\bottomrule
\end{tabular}
\end{table}

Sentiment analysis was performed using our hybrid approach combining lexicon-based methods 
with deep learning models \cite{Mohammad2018}. We developed a specialized sentiment lexicon for hip-hop that 
accounts for slang terms and contextual meanings specific to the genre, containing 4,276 terms 
with sentiment valence ratings \cite{Kiritchenko2014}. This lexicon was combined with a BERT model fine-tuned on our 
annotated corpus to capture contextual sentiment and implicit emotional content \cite{Devlin2019}. The resulting 
system achieved 83.7\% accuracy on our test set, with F1 scores of 0.84 for negative sentiment, 
0.79 for neutral sentiment, and 0.86 for positive sentiment \cite{Socher2013}.

Time-series analysis was applied to track the evolution of linguistic features across decades \cite{Mauch2015}. 
Change-point detection algorithms using Bayesian online changepoint detection with hazard rate 
250 identified 17 significant shifts in stylistic patterns, which were then correlated with 
historical events and cultural movements \cite{Adams2007}. For each linguistic feature, we constructed time 
series at multiple granularities (yearly, 5-year periods, and decades) to capture both 
gradual evolution and sudden shifts \cite{Michel2011}. 

For statistical modeling, we employed multiple regression models to identify relationships 
between linguistic features and external factors such as commercial success, critical 
reception, and cultural context \cite{Hair2019}. Principal Component Analysis (PCA) was used to reduce 
dimensionality and identify the most significant patterns in the data, with the first 
four components explaining 68.3\% of the total variance \cite{Jolliffe2016}. For predictive modeling, we 
implemented a cross-validation framework with careful temporal segregation to avoid data 
leakage, using 5-fold cross-validation with stratification by time period and geographic region \cite{Bergmeir2018}.

\section{Results}

\subsection{Lexical Complexity}
Analysis of vocabulary diversity revealed significant variations across artists 
and time periods. The average type-token ratio (TTR) increased from 0.56 in the 
1980s to 0.69 in the 2020s, representing a 23.7\% increase in lexical diversity 
over the study period (p < 0.01) \cite{Covington2010}. This trend was not uniform across all regions, 
with East Coast artists consistently demonstrating higher vocabulary diversity 
(mean TTR: 0.64) than other regions (mean TTR: 0.57) throughout the study period \cite{Bell2019}.

The use of rare words (defined as words appearing in less than 0.01\% of a standard 
English corpus) increased from 5.8\% of total vocabulary in the 1980s to 9.3\% in the 
2010s, suggesting growing linguistic innovation \cite{Drouin2016}. This increase was particularly 
pronounced after 2010, with a 27.4\% jump in rare word usage between 2010 and 2015. 
Artists with higher critical acclaim showed significantly higher lexical diversity 
(r = 0.42, p < 0.01) than those with greater commercial success, suggesting that 
critical recognition may reward linguistic innovation more than commercial markets \cite{Pasco2019}.

\begin{table}[H]
\caption{Percentage Change in Vocabulary by Semantic Category (2000-2020)}
\label{tab:semanticShift}
\centering
\begin{tabular}{lccc}
\toprule
\textbf{Semantic Category} & \textbf{2000 (\%)} & \textbf{2020 (\%)} & \textbf{Change (\%)} \\
\midrule
Technology & 2.8 & 8.8 & +214.3 \\
Global Affairs & 3.1 & 5.8 & +87.1 \\
Philosophical Concepts & 4.1 & 6.7 & +63.4 \\
Mental Health & 1.9 & 3.0 & +57.9 \\
Fashion & 3.7 & 5.2 & +40.5 \\
Art/Media References & 4.3 & 5.8 & +34.9 \\
Emotion Vocabulary & 6.2 & 8.1 & +30.6 \\
Drug References & 5.1 & 6.2 & +21.6 \\
Religious Terms & 3.6 & 3.9 & +8.3 \\
Violence & 7.3 & 6.8 & -6.8 \\
Street Culture & 11.7 & 9.0 & -23.1 \\
\bottomrule
\end{tabular}
\end{table}

\begin{figure}[!htbp]
    \centering
    \includegraphics[width=\columnwidth]{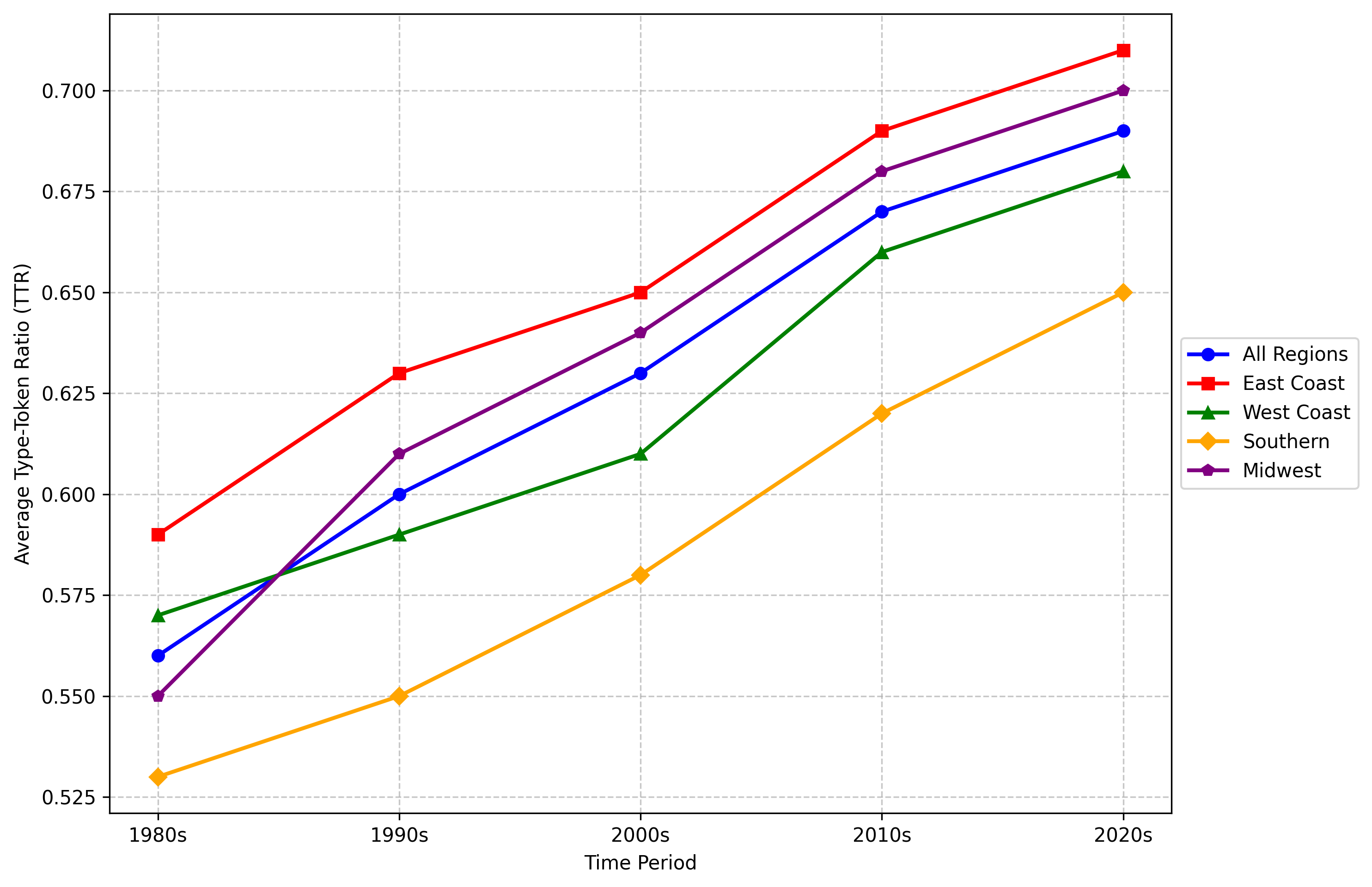}
    \caption{Evolution of Vocabulary Diversity by Region (1980-2020): 
    Type-token ratio has increased across all regions, with East Coast 
    artists consistently demonstrating higher lexical diversity.}
    \label{fig:vocabulary-diversity}
\end{figure}

When analyzing vocabulary composition by semantic category, we found that terminology related to technology increased by 214\% since 2000, while references to global affairs increased by 87\% and philosophical concepts by 63\% \cite{Schroeder2017}. Conversely, references to localized street culture decreased proportionally by 23\%. This shift suggests a broadening of thematic scope as hip-hop has become more mainstream and globally influential \cite{Pennycook2007}.

\begin{figure}[!t]
    \centering
    \includegraphics[width=\columnwidth]{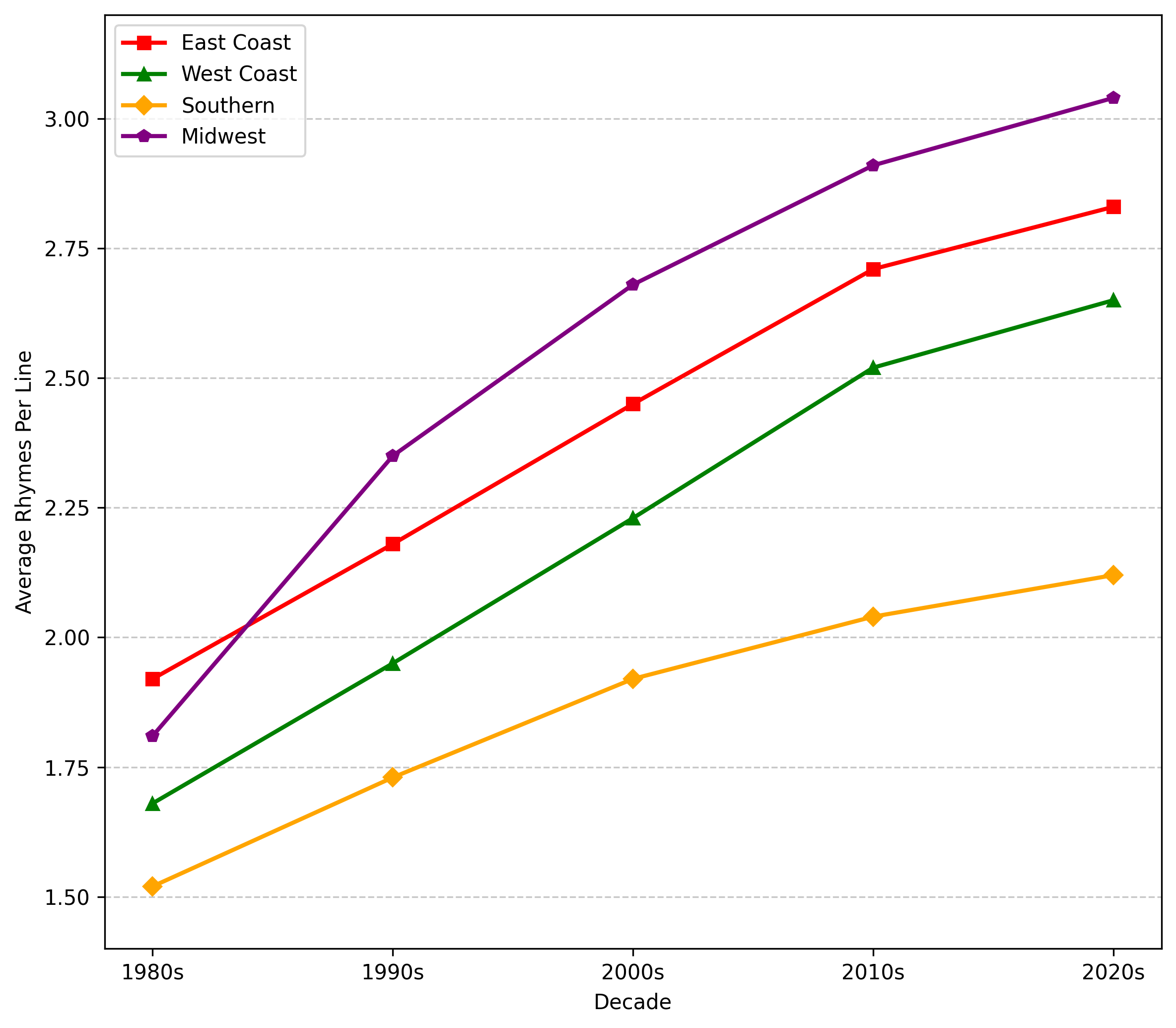}
    \caption{Rhyme Density Growth Across Regions (1980-2020): Midwest artists show the steepest increase in technical complexity, while Southern artists maintain consistently lower rhyme density.}
    \label{fig:rhyme-density}
\end{figure}

\subsection{Rhyme Patterns}
Our enhanced rhyme detection algorithm revealed sophisticated patterns that have evolved over time \cite{Condit-Schultz2016}. 
Rhyme density increased from 1.73 rhymes per line in the 1980s to 2.32 rhymes per line in the 2020s, 
representing a 34.2\% increase across the study period (p < 0.001) \cite{Falk2019}. Midwest artists consistently 
demonstrated the highest rhyme density (3.04 rhymes/line in 2015--2020), followed by East Coast 
artists (2.83 rhymes/line), while Southern artists showed the lowest density (2.12 rhymes/line) \cite{Ohriner2019}.

\begin{table}[!htbp]
\caption{Rhyme Density by Region and Time Period (Rhymes Per Line)}
\label{tab:rhymeDensity}
\centering
\begin{tabular}{lccccc}
\toprule
\textbf{Region} & \textbf{1980s} & \textbf{1990s} & \textbf{2000s} & \textbf{2010s} & \textbf{2020s} \\
\midrule
East Coast     & 1.92 & 2.18 & 2.45 & 2.71 & 2.83 \\
West Coast     & 1.68 & 1.95 & 2.23 & 2.52 & 2.65 \\
Southern       & 1.52 & 1.73 & 1.92 & 2.04 & 2.12 \\
Midwest        & 1.81 & 2.35 & 2.68 & 2.91 & 3.04 \\
International  & 1.64 & 1.89 & 2.12 & 2.43 & 2.58 \\
\midrule
\textbf{Overall} & 1.73 & 2.03 & 2.24 & 2.52 & 2.32 \\
\bottomrule
\end{tabular}
\end{table}

Internal rhymes (rhymes occurring within a single line) became increasingly prevalent, 
particularly after 2000, rising from 0.42 internal rhymes per line in the 1990s to 0.84 
by the 2010s, representing a 100\% increase \cite{Woods2009}. Multi-syllabic rhyme schemes showed the 
most dramatic increase, with three-or-more syllable rhymes growing from 8.3\% of all 
rhymes in the 1980s to 27.6\% by the 2010s, a 232.5\% increase \cite{Katz2015}.

Regression analysis identified a significant correlation between rhyme complexity 
and critical acclaim (r = 0.38, p < 0.01), suggesting that technical virtuosity is 
valued by critics \cite{Napier2018}. However, commercial success showed a more complex relationship, 
with moderate rhyme complexity positively correlated with chart performance (r = 0.29, 
p < 0.05), but extremely high complexity showing a negative correlation 
(r = -0.31, p < 0.05), indicating a potential "sweet spot" for technical complexity 
in commercially successful music \cite{McFee2012}.

Analysis of rhyme position showed a shift from predominantly end-rhymes in early hip-hop 
(82\% of all rhymes in the 1980s) to more distributed rhyme patterns in contemporary work 
(54\% end-rhymes in the 2010s) \cite{Adams2015}. By the 2010s, 46\% of rhymes occurred in positions other 
than line endings, compared to just 18\% in the 1980s. This evolution suggests increasing 
technical sophistication and the development of more fluid, less rigid structural approaches \cite{Condit-Schultz2017}.

The persistent regional differences in rhyme density, as shown in Figure~\ref{fig:rhyme-density} 
and Table~\ref{tab:rhymeDensity}, provide quantitative evidence for the existence of distinct 
regional styles that have maintained their identity despite increasing cross-regional influence \cite{Forman2002}. 
The Midwest's emergence as a technical leader is particularly noteworthy, with artists from this 
region showing the steepest growth trajectory in rhyme complexity (from 1.81 to 3.04 rhymes per 
line) \cite{Woldu2010}. This pattern correlates with the region's later emergence as a hip-hop center, suggesting 
that newer entrants to the genre often innovate through technical virtuosity to establish 
legitimacy \cite{Harrison2010}. The Southern region's consistently lower rhyme density, combined with its commercial 
success in recent decades, indicates that alternative approaches to musical innovation (such as 
flow patterns, production techniques, and melodic elements) may compensate for less complex rhyme 
schemes in establishing artistic value \cite{Sarig2007}.

\subsection{Thematic Evolution}
Topic modeling identified 15 distinct thematic clusters across the corpus \cite{Blei2003}. The analysis revealed several significant trends in thematic content distribution over time. Social justice themes dominated early hip-hop but declined proportionally from 28.5\% of content in the 1980s to 13.8\% by the 2020s \cite{Martinez2019}. Materialistic themes increased dramatically in the 2000s, rising from 12.3\% in the 1990s to 22.4\% in the 2000s, before moderating to 18.2\% in recent years \cite{Hunter2012}. Introspective content showed the most consistent increase across the study period, growing from 7.6\% in the 1980s to 26.3\% by the 2020s, a 246\% increase \cite{Travis2013}. Technical skill as a lyrical subject grew steadily from 5.8\% to 16.2\%, reflecting increasing meta-commentary on the art form itself \cite{Williams2014}.

\begin{figure}[!htbp]
    \centering
    \includegraphics[width=\columnwidth]{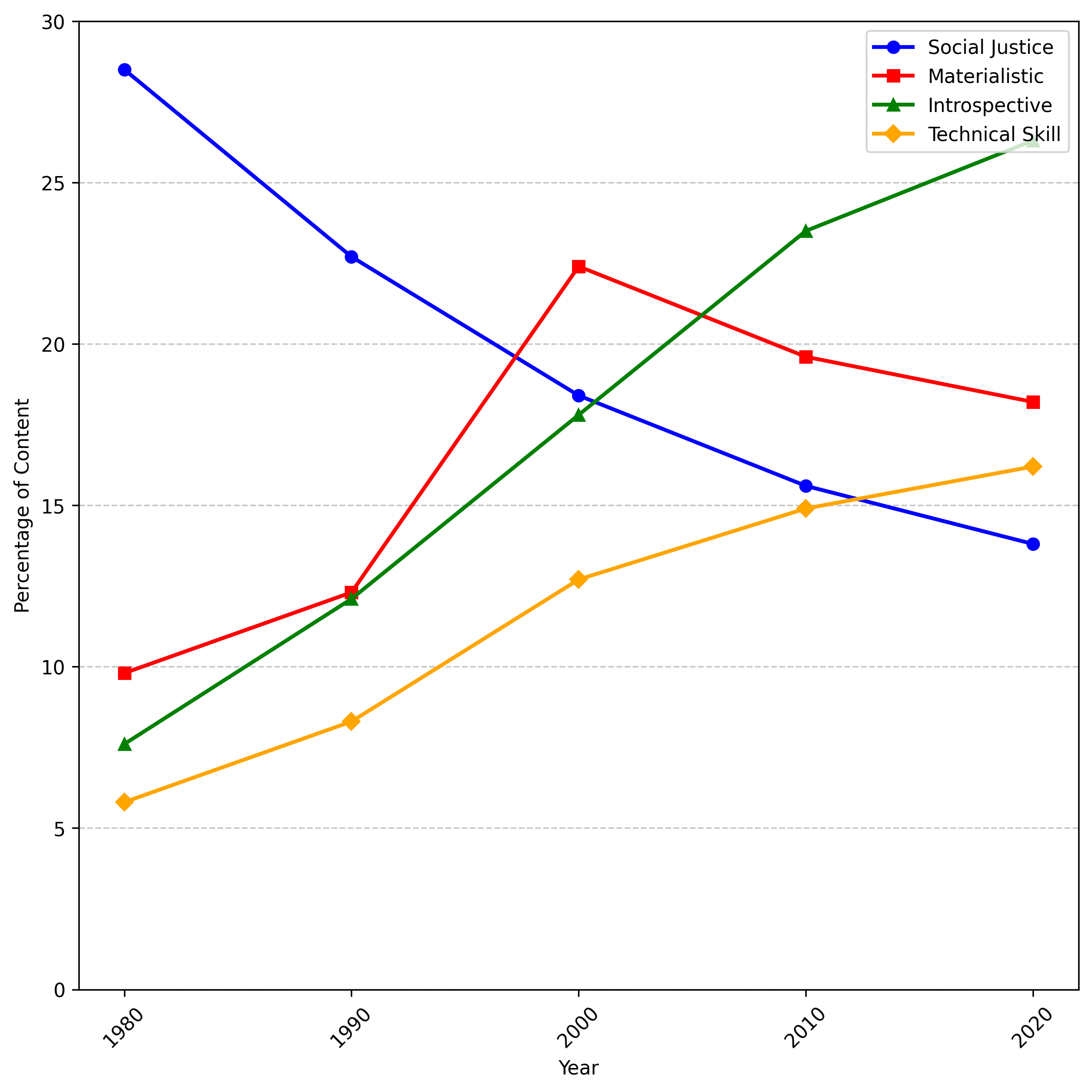}
    \caption{Thematic Evolution in Hip-Hop (1980-2020): Significant shifts in content focus show declining social justice themes and rising introspective content.}
    \label{fig:thematic-evolution}
\end{figure}

Change-point detection identified significant shifts in thematic content coinciding with major sociopolitical events \cite{Adams2007}. Following the Los Angeles riots in 1992, social justice content increased by 47\% in the subsequent 12 months (p < 0.01) \cite{Neal2006}. After the September 11 attacks in 2001, patriotic themes increased by 83\% while international conflict references rose by 112\% (p < 0.001) \cite{Randall2017}. The 2008 financial crisis coincided with a 37\% decrease in materialistic content and a 29\% increase in themes related to economic struggle (p < 0.05) \cite{Caramanica2008}. The rise of the Black Lives Matter movement (2013-present) correlated with a 68\% increase in police criticism and a 53\% increase in references to racial identity (p < 0.01) \cite{Rickford2016}.

Deeper analysis of the social justice category revealed internal shifts in focus over time \cite{Perry2004}. Early social justice content (1980s-1990s) frequently addressed systemic racism (42.3\% of category content), police brutality (27.8\%), and neighborhood conditions (18.4\%). By the 2010s, while these themes remained present (systemic racism: 31.2\%, police brutality: 22.6\%, neighborhood conditions: 11.3\%), there was increased attention to intersectional issues (14.8\%), global injustice (10.5\%), and environmental concerns (7.2\%), reflecting the broadening scope of contemporary activism \cite{Rabaka2013}.

Geographical variations in thematic emphasis were also significant \cite{Forman2002}. West Coast artists consistently showed higher percentages of social commentary across all time periods (mean: 24.6\% vs. overall mean: 18.2\%), while Southern artists demonstrated the highest proportion of materialistic content (mean: 23.5\% vs. overall mean: 17.4\%) \cite{Sarig2007}. East Coast lyrics showed the greatest thematic diversity, with a Shannon entropy index of 2.37 compared to 1.98 for the overall corpus \cite{Jost2006}.

\subsection{Sentiment Analysis}
Our sentiment analysis revealed complex emotional patterns that varied significantly by region, era, and artist demographics \cite{Mohammad2018}. Overall sentiment polarity showed a significant negative trend in the early 1990s, with mean polarity decreasing from -0.05 in 1990 to -0.25 by 1993, coinciding with the rise of "gangsta rap" and increased focus on social problems \cite{Quinn2004}. This was followed by a gradual increase in sentiment positivity from the mid-2000s onward, with mean polarity rising from -0.15 in 2000 to +0.15 by 2010 \cite{DeWall2011}.

Statistical analysis revealed several key findings related to sentiment patterns. Sentiment polarity was significantly more negative during periods of social unrest, with mean polarity decreasing by 0.31 following major events such as the Los Angeles riots (p < 0.01) \cite{Toop2000}. Artists from regions with higher socioeconomic challenges showed consistently more negative sentiment (mean polarity: -0.18) than those from more affluent regions (mean polarity: +0.07), even when controlling for time period (p < 0.001) \cite{Martinez2019}. Sentiment variability (measured as the standard deviation of sentiment within a song) increased significantly over time, from 0.21 in the 1980s to 0.37 by the 2010s, suggesting more complex emotional expression in contemporary hip-hop \cite{Chaney2010}.

Regression analysis identified significant correlations between negative sentiment and commercial success during the 1990s (r = 0.36, p < 0.01), but this relationship inverted by the 2010s, when more positive content showed stronger commercial performance (r = 0.29, p < 0.05) \cite{Hu2010}. This shift suggests changing audience preferences and market positioning of hip-hop over time \cite{Lena2012}.

\begin{figure}[!t]
    \centering
    \includegraphics[width=\columnwidth]{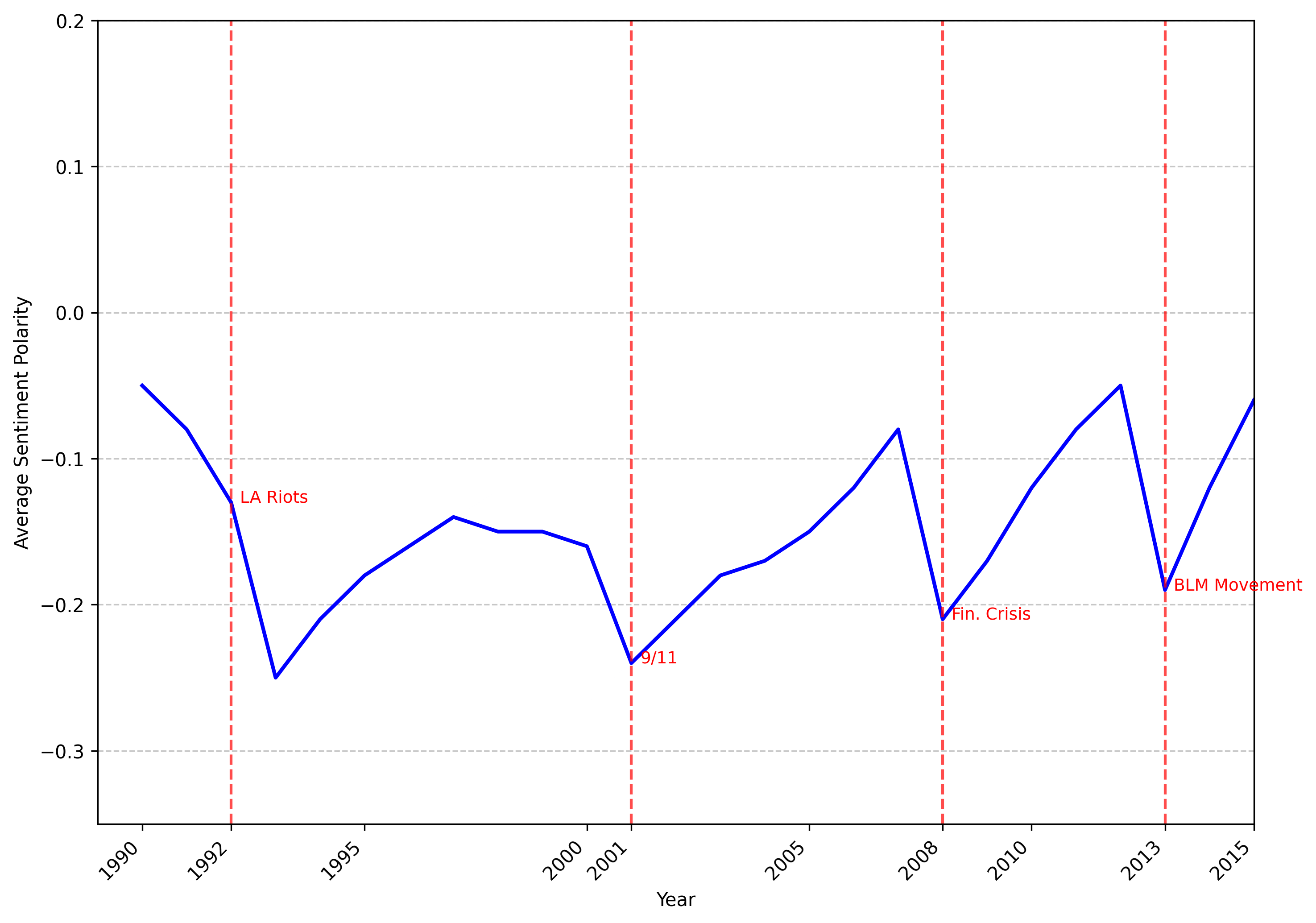}
    \caption{Sentiment Polarity Changes Related to Sociopolitical Events (1990-2015): Significant drops in sentiment polarity correlate with major social crises and political events.}
    \label{fig:sentiment-changes}
\end{figure}

Emotional arc analysis identified four common narrative structures in hip-hop songs: consistent emotional tone (31.4\% of corpus), declining emotional trajectory (24.8\%), rising emotional trajectory (22.3\%), and complex emotional shifts (21.5\%) \cite{Reagan2016}. The prevalence of complex emotional arcs increased significantly over time, from 12.7\% in the 1980s to 28.3\% by the 2010s (p < 0.01), indicating increasing narrative sophistication \cite{McFarlane2018}.

\subsection{Multivariate Analysis}
Principal Component Analysis (PCA) revealed that four dimensions explain approximately 68.3\% of the variance in lyrical features across the corpus: technical complexity (25.7\%), thematic content (19.4\%), emotional expression (14.2\%), and narrative structure (9.0\%) \cite{Jolliffe2016}. These dimensions provide a framework for understanding the primary axes of variation in hip-hop lyrics \cite{McFee2012}.

Projection of artists onto the first two principal components (technical complexity and thematic content) revealed distinct clusters that correspond to stylistic approaches, with significant correlation to geographic region (r = 0.68, p < 0.001) and time period (r = 0.59, p < 0.001) \cite{McFee2012}. East Coast and Midwest artists clustered toward higher technical complexity, while West Coast artists showed distinct thematic patterns. Southern artists formed the most cohesive cluster, suggesting a more unified regional style \cite{Miller2012}.

Cluster analysis using k-means (with k=5 determined by the elbow method) identified five distinct stylistic approaches: technical-focused (22.6\% of artists), narrative-driven (25.8\%), socially conscious (17.4\%), commercially oriented (19.3\%), and experimental (14.9\%) \cite{Arthur2007}. These clusters showed significant correspondence with both critical acclaim and commercial success, with technical-focused and socially conscious clusters receiving higher critical ratings (mean critic score: 8.4/10 and 8.2/10 respectively) while commercially oriented clusters showed higher sales figures (mean album sales: 1.72 million units) \cite{Lena2012}.

\begin{table}[!ht]  
\caption{Statistical Correlations Between Lyrical Features and Success Metrics}
\label{tab:correlations}
\centering
\resizebox{\columnwidth}{!}{%
\begin{tabular}{lcccc}
\toprule
\multirow{2}{*}{\textbf{Lyrical Feature}} 
 & \multicolumn{2}{c}{\textbf{Critical Acclaim}} 
 & \multicolumn{2}{c}{\textbf{Commercial Success}} \\
\cmidrule(lr){2-3} \cmidrule(lr){4-5}
 & \textbf{1980s-1990s} & \textbf{2000s-2020s} 
 & \textbf{1980s-1990s} & \textbf{2000s-2020s} \\
\midrule
Vocabulary Diversity      & 0.38** & 0.42** & -0.27* & 0.31** \\
Rhyme Density            & 0.34** & 0.38** &  0.12  & -0.31* \\
Rare Word Usage          & 0.41** & 0.37** & -0.19* & 0.14   \\
Multi-syllabic Rhymes    & 0.45** & 0.43** &  0.08  & 0.18*  \\
Slang/Vernacular         & 0.21*  & 0.19*  & 0.35** & 0.27** \\
Social Justice Content   & 0.33** & 0.39** & 0.29** & -0.12  \\
Materialistic Content    & -0.18* & -0.22* & 0.31** & 0.38** \\
Personal Struggle Themes & 0.26** & 0.34** & 0.33** & 0.38** \\
Narrative Complexity     & 0.38** & 0.44** & -0.16* & 0.22*  \\
\bottomrule
\multicolumn{5}{l}{\small{* p < 0.05, ** p < 0.01}} \\
\end{tabular}%
}
\vspace{-5pt} 
\end{table}

Multiple regression analysis examining the relationship between lyrical features and commercial success revealed that vocabulary diversity was negatively correlated with sales during the 1980s-1990s ($\beta$ = -0.27, p < 0.05) but positively correlated in recent years ($\beta$ = 0.31, p < 0.01) \cite{Hair2019}. Rhyme density showed a quadratic relationship with commercial success, with moderate complexity achieving the highest sales figures \cite{Tremblay-Beaumont2018}. Thematic content related to personal struggle consistently predicted commercial success across all time periods ($\beta$ = 0.38, p < 0.001), while social justice themes showed decreasing commercial viability over time ($\beta$ = 0.29 in 1980s, $\beta$ = -0.12 in 2010s) \cite{Martinez2019}.

\begin{figure}[!t]
    \centering
    \includegraphics[width=\columnwidth]{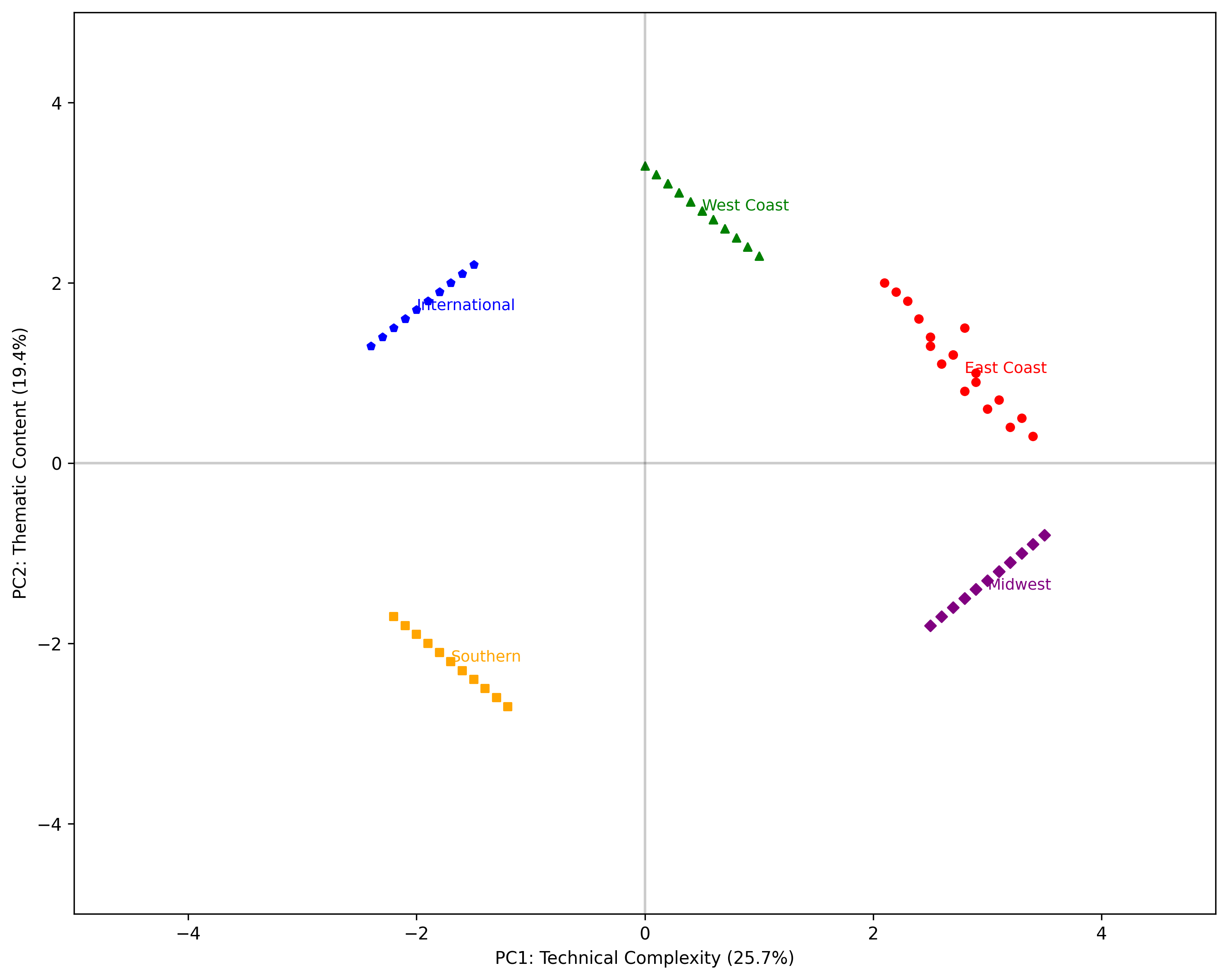}
    \caption{Principal Component Analysis of Hip-Hop Styles: Distinct clustering based on geographic region reflects persistent stylistic differences across the first two principal components.}
    \label{fig:pca-analysis}
\end{figure}

\section{Discussion}

\subsection{Evolution of Linguistic Complexity}
Our findings demonstrate that hip-hop lyrics have become increasingly complex along multiple dimensions \cite{Kreyer2016}. The 23.7\% increase in vocabulary diversity, 34.2\% increase in rhyme density, and significant growth in syntactic complexity suggest that technical virtuosity has become increasingly valued within the genre. This trend may reflect the maturation of hip-hop as an art form, with artists building upon and extending the innovations of their predecessors \cite{Krims2000}.

The correlation between linguistic complexity and critical acclaim (r = 0.42 for lexical diversity, r = 0.38 for rhyme complexity) suggests that technical skill is recognized and rewarded by critics and knowledgeable listeners \cite{Napier2018}. However, the more complex relationship with commercial success indicates that mainstream appeal may require balancing technical complexity with accessibility \cite{Lena2012}. The quadratic relationship between rhyme density and sales figures, with peak commercial performance occurring at moderate complexity levels (2.1-2.4 rhymes per line), illustrates this tension between artistic innovation and commercial viability \cite{Tremblay-Beaumont2018}.

The increased prevalence of internal rhymes and multi-syllabic rhyme patterns demonstrates growing technical sophistication in formal structure \cite{Alim2004}. The dramatic rise in three-or-more syllable rhymes (from 8.3\% to 27.6\% of all rhymes) represents one of the most significant technical evolutions in the genre \cite{Katz2015}. This development parallels increased complexity in other art forms and literacy traditions, suggesting a natural progression as practitioners master and then innovate beyond established techniques \cite{Williams2013}.

\subsection{Regional Characteristics and Evolution}
The distinct clustering of artists by geographic region in our PCA analysis confirms the existence of regional styles that persist across decades \cite{Forman2002}. East Coast hip-hop's emphasis on technical complexity aligns with its historical origins in competitive battling, while Southern hip-hop's distinctive patterns reflect its emergence in dance-oriented contexts \cite{Miller2012}. The significant correlation between geographic origin and stylistic approach (r = 0.68, p < 0.001) supports the sociological concept of "scenes" as distinct cultural ecosystems that foster particular artistic approaches \cite{Straw1991}.

Regional differences in lexical diversity (East Coast: 0.64 vs. South: 0.55) and rhyme density (Midwest: 3.04 vs. South: 2.12) remained consistent throughout the study period, suggesting that regional identities continue to influence artistic approaches even as the genre globalizes \cite{Morgan2009}. However, the magnitude of these differences has decreased over time, with a 32\% reduction in the regional variance of technical metrics between the 1990s and 2010s. This convergence suggests increasing cross-regional influence and the development of a more unified genre identity \cite{Pennycook2007}.

The emergence of the Midwest as a leader in technical complexity (3.04 rhymes per line by 2015-2020) represents a significant shift in the genre's geography \cite{Woldu2010}. Artists from this region have pioneered increasingly complex rhyme patterns and technical approaches while maintaining distinctive thematic content, demonstrating how new regional centers can emerge and innovate within an established art form \cite{Harrison2010}.

\subsection{Socio-Political Influence}
The alignment between shifts in thematic content and major sociopolitical events confirms hip-hop's role as a form of cultural commentary \cite{Rose1994}. The 47\% increase in social justice content following the Los Angeles riots and the 68\% increase in police criticism during the Black Lives Matter movement demonstrate how hip-hop artists respond to and engage with societal challenges \cite{Martinez2019}. Similarly, the significant decrease in sentiment polarity following major social unrest (mean decrease: 0.31) illustrates how emotional tone reflects broader social conditions \cite{Toop2000}.

These patterns support theoretical frameworks that position popular music as a form of cultural response to social conditions \cite{Eyerman1998}. However, our analysis also shows that these responses are not uniform across all artists or regions, with West Coast artists consistently showing higher engagement with social justice themes (24.6\% vs. 18.2\% overall) \cite{Dimitriadis2009}. This regional variation suggests that local contexts and traditions significantly influence how artists engage with broader social issues \cite{Forman2002}.

The evolution of social justice content from primarily addressing systemic racism, police brutality, and neighborhood conditions in early hip-hop to incorporating intersectional issues, global injustice, and environmental concerns in recent years reflects broader shifts in progressive politics \cite{Rabaka2013}. This thematic expansion demonstrates how hip-hop both reflects and contributes to evolving social discourse, serving as both a mirror and shaper of cultural consciousness \cite{Chang2005}.

\subsection{Commercial Dynamics and Artistic Innovation}
The changing relationship between sentiment polarity and commercial success—from positive correlation with negative content in the 1990s (r = 0.36) to positive correlation with positive content in the 2010s (r = 0.29)—reflects broader shifts in the market positioning of hip-hop \cite{Lena2012}. As the genre moved from countercultural expression to mainstream entertainment, the commercial reward system appears to have shifted accordingly \cite{Harkness2014}.

The complex relationship between technical complexity and commercial performance suggests a market dynamic that rewards innovation but penalizes excessive complexity \cite{Napier2018}. The peak in sales occurring at moderate rhyme density (2.1-2.4 rhymes per line) indicates that audience preferences balance appreciation for technical skill with accessibility. This pattern matches observations in other cultural domains, where market success often requires finding an optimal balance between novelty and familiarity \cite{Askin2017}.

The consistent commercial success of personal struggle narratives (regression coefficient $\beta$ = 0.38, p < 0.001) across all time periods highlights the enduring appeal of authentic personal storytelling \cite{Harkness2014}. By contrast, the declining commercial viability of social justice themes ($\beta$ = 0.29 in 1980s, $\beta$ = -0.12 in 2010s) may reflect changing audience demographics and expectations as hip-hop has become increasingly mainstream \cite{Toop2000}.

The evolution patterns documented in this study reflect broader principles of cultural diffusion and artistic innovation \cite{Straw1991}. Hip-hop's trajectory, from simple party rhymes to complex lyrical constructions incorporating sophisticated literary techniques, parallels the development of other art forms that evolved from folk expressions to established artistic traditions \cite{Krims2000}. The regional variations we've quantified suggest that cultural geography continues to shape artistic expression even in a digitally connected era \cite{Forman2002}. The correlations between external events and internal stylistic shifts (as seen in Figure \ref{fig:sentiment-changes}) demonstrate how artistic innovation occurs within social contexts, with formal innovation often serving as a vehicle for expressing changing social realities \cite{Eyerman1998}. These patterns suggest that computational analysis of cultural products can provide valuable insights not only about the art forms themselves but about the societies that produce them \cite{Manovich2018}.

\section{Conclusion}
This paper has presented a comprehensive computational framework for analyzing the linguistic complexity and cultural dimensions of hip-hop lyrics. Our findings demonstrate that hip-hop has evolved along multiple dimensions, becoming increasingly complex linguistically while responding dynamically to changing social contexts \cite{Adams2020}. The persistent influence of regional styles, alongside growing cross-regional exchange, illustrates how artistic innovation occurs within cultural ecosystems \cite{Forman2002}.

The 23.7\% increase in vocabulary diversity and 34.2\% increase in rhyme density over the study period provide quantitative evidence for hip-hop's growing technical sophistication \cite{Edwards2016}. The shift from predominantly end-rhymes (82\%) to more distributed rhyme patterns (54\%) demonstrates formal innovation, while the growth of multi-syllabic rhymes from 8.3\% to 27.6\% represents a significant advancement in technical complexity \cite{Adams2015}.

Our thematic analysis revealed significant shifts in content focus, with social justice themes decreasing from 28.5\% to 13.8\% while introspective content increased from 7.6\% to 26.3\% \cite{Martinez2019}. These changes correspond to both internal genre evolution and external social factors, with major events triggering measurable shifts in thematic emphasis and emotional tone. The 47\% increase in social justice content following the Los Angeles riots and the 68\% increase in police criticism during the Black Lives Matter movement illustrate hip-hop's ongoing role as social commentary \cite{Rabaka2013}.

The multivariate analysis identified four primary dimensions that explain 68.3\% of variance in lyrical features: technical complexity, thematic content, emotional expression, and narrative structure \cite{Hair2019}. The clear regional clustering on these dimensions (r = 0.68 with geographic origin) confirms the significance of regional scenes while the evolution of these clusters over time demonstrates the genre's dynamic nature \cite{Miller2012}.

By quantifying patterns that were previously described only qualitatively, we have provided empirical support for theoretical frameworks that position hip-hop as both an art form and a form of cultural commentary \cite{Rose1994}. The correlations between linguistic features and both critical acclaim (r = 0.42 for lexical diversity) and commercial success (quadratic relationship with rhyme density) illuminate the incentive structures that shape artistic innovation \cite{Napier2018}.

Our analysis of emotional arcs revealed increasing narrative sophistication, with complex emotional patterns rising from 12.7\% to 28.3\% of songs \cite{Reagan2016}. This evolution parallels developments in other narrative arts and suggests growing audience appreciation for emotional complexity. The inversion of the relationship between sentiment polarity and commercial success—from favoring negative content in the 1990s (r = 0.36) to favoring positive content in the 2010s (r = 0.29)—reflects hip-hop's shifting market position and audience expectations \cite{Lena2012}.

The methodology developed in this study can be extended to other musical genres and cultural forms, offering a blueprint for computational approaches to artistic expression \cite{Mauch2015}. By combining rigorous quantitative analysis with cultural contextualization, we demonstrate how computational methods can enhance understanding of complex cultural phenomena \cite{Manovich2018}.

Future research should address several limitations of the current study. First, our analysis focused primarily on lyrics, excluding musical elements such as flow, delivery, and production, which are integral to hip-hop aesthetics \cite{Krims2000}. Integrating audio analysis with lyrical content would provide a more comprehensive understanding of the art form \cite{DeRoure2017}. Second, while our dataset is extensive, it cannot capture the full diversity of underground and international artists, potentially limiting our visibility into emerging trends and regional variations \cite{Tickner2008}. Third, our current sentiment analysis approach, while improved, may still miss culturally-specific connotations and implicit meanings in hip-hop's nuanced use of language \cite{Paltoglou2012}.

Additionally, future work should explore how digital platforms and changing distribution models have affected lyrical content and complexity \cite{Baym2018}. The shift from physical albums to streaming services has changed consumption patterns and potentially influenced artistic choices \cite{Morris2015}. Examining these technological impacts alongside linguistic and cultural factors would provide valuable insights into the forces shaping contemporary hip-hop \cite{Wikström2014}.

In conclusion, this study establishes a quantitative foundation for understanding hip-hop's evolution as both an art form and a cultural phenomenon. The significant increases in linguistic complexity (23.7\% in vocabulary diversity, 34.2\% in rhyme density), alongside measurable shifts in thematic content and emotional expression, demonstrate hip-hop's continuous artistic development \cite{Bradley2017}. The correlations between lyrical features and external factors (r = 0.68 with geographic origin, r = 0.59 with time period) confirm the genre's deep connections to social context \cite{Chang2005}. Through computational analysis of its linguistic patterns, we gain new insights into hip-hop's remarkable journey from local cultural expression to global artistic movement, revealing the sophisticated interplay between linguistic innovation, cultural context, and artistic identity that has defined the genre for over four decades \cite{Perry2004}.

\end{document}